\documentclass{article}
\usepackage{booktabs}
\usepackage[table]{xcolor}
\usepackage{graphicx}
\usepackage{amsmath}
\usepackage{amssymb}
\usepackage{bbm}
\usepackage{booktabs}
\usepackage{makecell}
\usepackage{enumitem}
\usepackage{wrapfig}
\usepackage{pgfplots}
\usepackage{caption}
\pgfplotsset{compat=1.18}
\PassOptionsToPackage{numbers,sort&compress}{natbib}

\newcommand{\safeset}{\mathcal{S}}

\newcommand{\constraintset}{\mathcal{C}}

\newcommand{\safeactionset}{\mathcal{U}_{\mathrm{safe}}}

\newcommand{\safeactionsetopt}{\mathcal{U}_{\mathrm{safe}}^*}

\newcommand{\pisafe}{\pi_{\mathrm{safe}}}
\newcommand{\piref}{\pi_{\mathrm{ref}}}

\newcommand{\hruleopen}{\vspace{0.5em}\hrule\vspace{0.25em} \noindent \;}
\newcommand{\hruleclose}{\hrule \vspace{0.25em}}
\newcommand{\traj}{\mathrm{x}}
\newcommand{\ctrl}{\mathrm{u}}
\newcommand{\R}{\mathbb{R}}
\renewcommand{\Pr}{\mathrm{Pr}}

\newcommand{\constraintfcn}{c}
\newcommand{\costfcn}{g}

\usepackage[preprint]{corl_2026} 

\title{ResSafe: Learning Safety Filtering with Residual Reinforcement Learning for Humanoids}

\author{
  \textbf{Gechen Qu}\textsuperscript{1} \quad
  \textbf{Tong Zhang}\textsuperscript{1} \quad
  \textbf{Bike Zhang}\textsuperscript{1} \quad
  \textbf{Yen-Jen Wang}\textsuperscript{1} \\[0.3em]
  \textbf{Koushil Sreenath}\textbf{\textsuperscript{1,\dag}} \quad
  \textbf{Claire Tomlin}\textbf{\textsuperscript{1,\dag}} \quad
  \textbf{Jason Jangho Choi}\textbf{\textsuperscript{2,\dag}} \\[0.5em]
  \textsuperscript{1}University of California, Berkeley \qquad
  \textsuperscript{2}University of California, Los Angeles \\[0.3em]
}

\date{}

\begin{document}
\maketitle
{\renewcommand{\thefootnote}{\fnsymbol{footnote}}
\footnotetext[2]{Equal advising.}}

{\let\thefootnote\relax\footnotetext{Correspondence: \texttt{qugch@berkeley.edu}. Website: \href{https://sciautonomy.github.io/ResSafe_Web/}{\texttt{sciautonomy.github.io/ResSafe\_Web}}}}

\vspace{-2.5em}
\begin{figure}[h]
\centering
\vspace{-0.5em}
\includegraphics[width=0.8\textwidth]{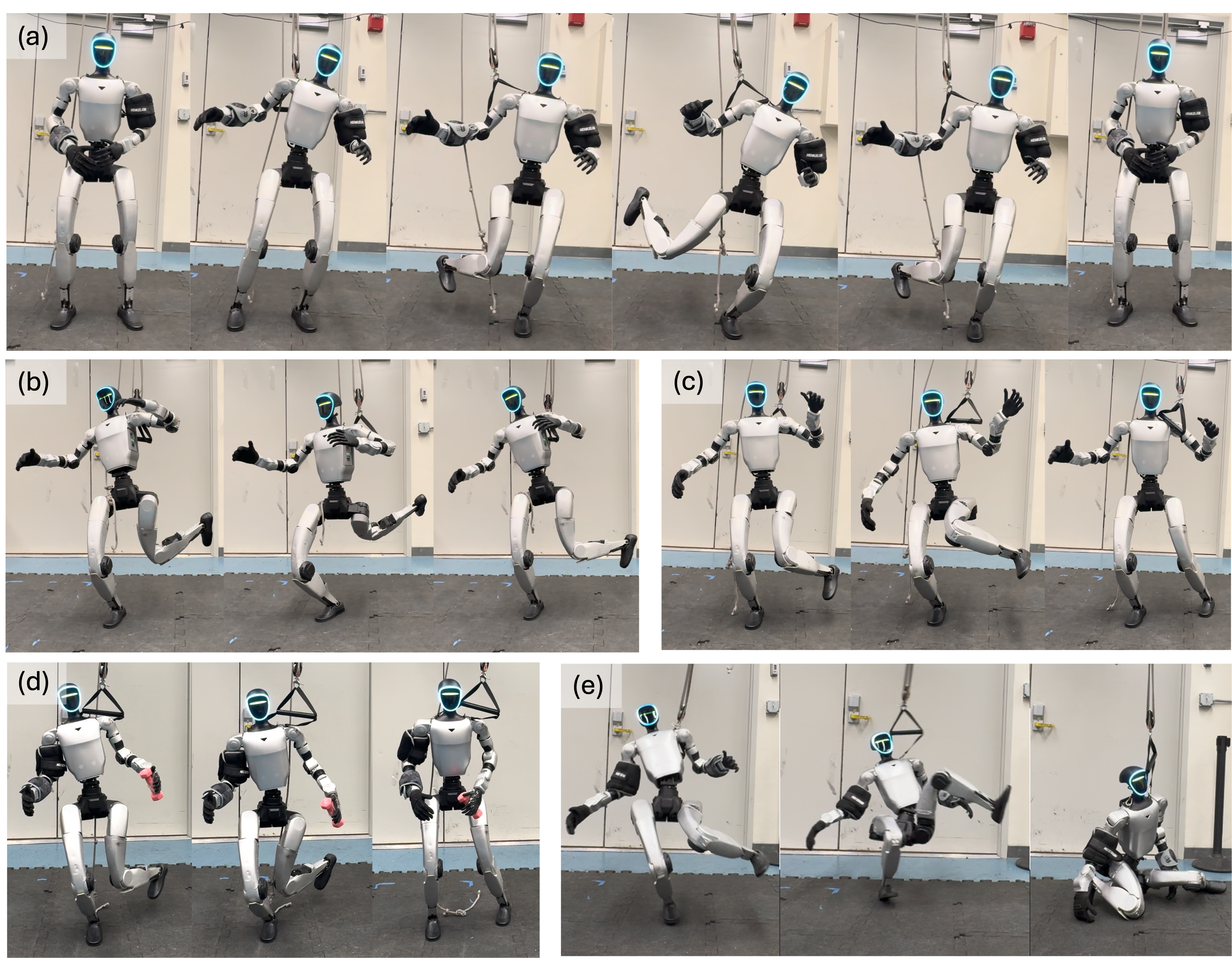}
\vspace{-0.5em}
 \caption{\small Humanoid balance tasks under various payloads. We introduce ResSafe, a performance-safety decoupling framework based on residual reinforcement learning which learns safety filtering. (a)-(d) Successful experiments of balance motions under our policy. (e) Failure case of the baseline policy. See more details in Section~\ref{sec:hardware} and Table~\ref{tab:real_world_results}.}

\label{fig:frontfigure}
\vspace{-1em}
\end{figure}

\begin{abstract}
    Safe control of humanoid robots remains challenging due to their high-dimensional dynamics, contact-rich interactions, and sensitivity to disturbances. Although reinforcement learning has enabled effective locomotion and motion tracking, learned policies can still generate unsafe actions that lead to instability or falls. In this work, we propose residual reinforcement learning as an implicit safety-filtering mechanism for safe humanoid control. Instead of relying on a single nominal policy to simultaneously balance performance, safety, and robustness, we decouple performance and safety. The nominal policy focuses solely on task performance, while a residual policy learns safety corrections. This decoupling leads to a better performance--safety Pareto trade-off and avoids the need for careful tuning of multiple competing reward terms within a single policy training. We show that the residual policy can act as an implicit safety filter.
    ~We evaluate our method in challenging humanoid balance tasks. Compared with the nominal reference policy and a learning-based safety filter baseline, residual reinforcement learning improves safety while still achieving competent performance.
\end{abstract}



\section{Introduction}

\begin{figure}[t]
\centering
\vspace{-0.5em}
\includegraphics[width=\textwidth]{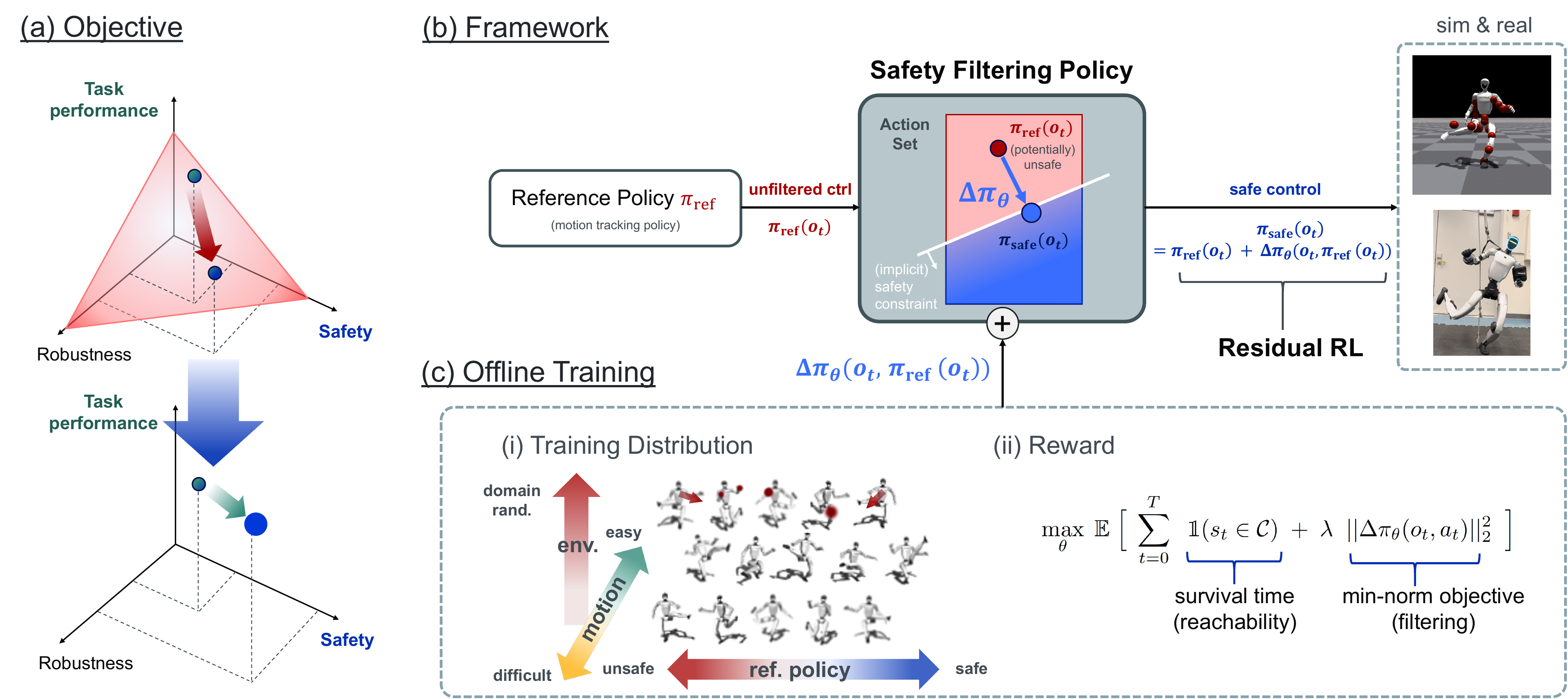}
\vspace{-1.25em}
 \caption{\small \textbf{ResSafe Overview.} \textbf{(a)} By decoupling the learning of performance and safety, we aim to achieve a better performance--safety Pareto frontier. \textbf{(b)} To achieve this, we propose a residual RL-based architecture in which the residual RL policy learns a safety correction, resembling the structure of safety filters \cite{ames2019control, l4dc}. \textbf{(c)} We train the residual RL safety filtering policy primarily through a maximum-survival-time reachability formulation. 
 To improve generalization across various reference actions, we expose the residual RL policy during training to varying levels of reference motion difficulty, reference policy safety, and domain randomization.
 }
\label{fig:method}
\vspace{-1em}
\end{figure}

Reinforcement learning (RL) for humanoid control inherently involves a multi-objective optimization problem, where nominal performance (e.g., agility or motion tracking) must be balanced against safety and robustness \cite{li2025reinforcement, zhang2025hub}. As illustrated in Fig.~\ref{fig:method}.(a), these objectives are often in conflict, leading to a Pareto trade-off. In practice, however, RL algorithms rely on gradient-based optimization and typically converge to locally optimal solutions, often resulting in a suboptimal Pareto frontier. Consequently, in current RL frameworks, \textit{improving robustness or safety}, e.g., through increased domain randomization or stronger penalties for failures, \textit{inevitably comes at the expense of degraded performance}. This limitation is also present in classical safe RL based on constrained Markov Decision Process (MDP) \cite{gu2024review, pmlr-v70-achiam17a}, where safety and performance are coupled through the Lagrangian.

In this work, we show that by \textit{decoupling the learning of performance and safety}, we can significantly improve the achievable performance--safety Pareto frontier of humanoid motion policies. This decoupling is inspired by human motor intelligence: when learning a new sport, humans do not relearn safety---such as maintaining balance and avoiding collisions---from scratch, but instead rely on reflexive motor skills developed through years of growth and interaction with the physical world. Motivated by this observation, our objective is to develop a generalizable safety filtering policy for humanoids that transfers across tasks without requiring retraining for safety from scratch. As an initial step toward this goal, we propose a framework that explicitly decouples the learning of performance and safety during training.

To instantiate this idea, we propose a residual reinforcement learning framework that separates nominal task optimization from safety correction. We first train a nominal motion policy optimized for task performance and then keep it fixed while training a residual safety policy. The final action is obtained by augmenting the nominal action with a learned safety residual correction (Fig.~\ref{fig:method}.(b)). 

Overall, our contributions can be summarized as follows: \vspace{-0.5em}
\begin{itemize}[itemsep=0em, leftmargin=1.5em]
    \item We propose ResSafe, a performance--safety decoupling framework for learning-based safety filtering and RL-based humanoid control, where a reference policy optimizes task performance and a separate residual policy learns safety corrections.

    \item We connect residual RL with control-theoretic safety filtering by showing that the learned residual policy behaves similarly to a min-norm safety filter correction, providing an implicit safety filtering mechanism without online optimization.

    \item We empirically demonstrate that the proposed decoupled learning framework achieves a superior performance--safety--robustness trade-off compared to single-policy training frameworks for humanoid extreme-balance tasks. We validate this framework in both simulation and hardware under random payloads, demonstrating improved robustness with small degradation in tracking performance, as well as generalization across different reference policy checkpoints.
    
\end{itemize}


\vspace{-0.5em}
\section{Related Works}
\vspace{-0.5em}

\textbf{Safe Control Theory and Safety Filters.} In classical safe control, safety is formulated as a state-constraint satisfaction problem, where the objective is to maintain the system within a control-invariant subset of the constraint set \cite{blanchini1999set}. Hamilton--Jacobi (HJ) reachability \cite{bansal} and control barrier functions (CBFs) \cite{ames2019control} address this problem through safety certificates that characterize the invariant set and safe control actions. These certificates can further be used to monitor and modify potentially unsafe reference actions through a safety filter \cite{Wabersich2023, hsu2023safety}. However, existing safety methods typically rely on explicit dynamics models and suffer from poor scalability with state dimension, limiting their applicability to systems with high-dimensional and complex dynamics such as humanoid robots.

\textbf{Safe Reinforcement Learning.}
Various safe RL algorithms proposed in \cite{pmlr-v70-achiam17a, garcia2015comprehensive, ray2019benchmarking, srinivasan2020learning, thananjeyan2021recovery, wagener2021safe, gu2024review} construe safety violations as an \textit{accumulation} of safety-relevant cost, referred to as the constrained MDP problem \cite{Altman1998}. Although such a formulation is applicable in certain safety settings, it does not adequately capture instantaneous constraint violations such as collisions or falls. Other formulations rooted in safe control theory are proposed, for example, Lyapunov theory in \cite{chow2018lyapunov, li2026clf}, and CBFs and HJ reachability in \cite{fisac2019bridging, huh2020safe, hsu2023isaacs, so2024train, li2025certifiable, yang2025cbf, oh2025safety, li2025verifiable,  kim2026deep, sharpless2026bellman}. However, these approaches rely on specialized reachability-type  Bellman equations, distinct from Bellman equations for typical sum-over-time rewards.

\textbf{Safety and Extreme Balance for Humanoids.} Various works are proposed to ensure or improve the safety of humanoids \cite{khazoom2022humanoid, baek2026whole, sun2026sparksafeprotectiveassistive, yang2025cbf, yang2025shield, lee2026safetycriticalwholebodycontrolhumanoid}, or locomotion more broadly \cite{choi2020reinforcement, nguyengameplay, he2024agile, yun2025safe, contractionppo}. However, the majority focuses on collision avoidance and stable walking, often using CBFs synthesized from simplified models, with relatively limited attention devoted to whole-body balance as a safety constraint \cite{contractionppo, lee2026safetycriticalwholebodycontrolhumanoid}. 
The works in \cite{zhang2025hub, pan2025agility} leverage refined motion datasets to achieve challenging balance motions; however, safety is handled through carefully hand-designed and tuned reward terms within a single-policy training framework.

\textbf{Residual Reinforcement Learning.} Residual RL combines prior controllers with learned residual policies, allowing robots to retain structured control knowledge while compensating for modeling errors, disturbances, and task-specific discrepancies \cite{johannink2019residual}. 
Recent works have extended this paradigm to various robotic systems, including dexterous manipulation~\cite{li2025maniptrans}, \cite{huang2025efficient}, quadcopter flight~\cite{zhang2025proxfly}, \cite{ishihara2023improving}, and humanoid whole-body control \cite{zhao2025resmimic}. 

\vspace{-0.75em}
\section{Background \& Problem Formulation}
\vspace{-0.5em}

\textbf{Constrained Optimal Control Problem \& Safety Filter.} We begin with a safety-constrained optimal control problem for a general nonlinear dynamical system and introduce the safety-filtering framework used to achieve performance--safety decoupling. Consider a trajectory $\traj(\cdot)$ of a dynamical system, governed by \vspace{-0.25em}
\begin{equation}
    \dot{\traj}(t) = f(\traj(t), \ctrl(t)) \;\;\text{for}\; t \ge 0,\;\; \traj(0) = x,
    \label{eq:dynsys}
\end{equation}
where $x\in \R^n$ is the initial state and $\ctrl$ is a control input signal in time. We assume that the control input at each time must satisfy a compact control input bound, $\ctrl(t) \in U \subset \R^m$, and consider all possible measurable control signals that satisfy the bound as the set of valid control signals $\mathcal{U}$.

The safety constraint is defined by \textit{the constraint set} $\constraintset \subset \R^n$, such that $\traj(t) \in \constraintset$ for all $t \ge 0$. The set can be represented as a zero-superlevel set of some continuous function $c:\R^n \rightarrow \R$, $
\constraintset = \{x \mid \constraintfcn(x) \ge 0 \}
$, or it can also be represented by an indicator function $\mathbbm{1}(x \in \constraintset)$. Finally, the control objective for the policy learning is
\begin{subequations}
\label{eq:constrained-opt}
\begin{align}
\inf_{\ctrl(\cdot) \in \mathcal{U}} &\int_{0}^{\infty} e^{-\gamma t} \costfcn(\traj(t), \ctrl(t)) dt \label{eq:cop-obj}
\\
\text{s.t.} \quad 
& \traj(t) \in \constraintset \;\; \forall t \ge 0, \label{eq:cop-constraint}
\end{align}
\end{subequations}
where \eqref{eq:cop-obj} represents the task objective, with the running cost function $g$ and a discount factor $\gamma > 0$, and \eqref{eq:cop-constraint} represents the safety constraint. We next decouple Problem \eqref{eq:constrained-opt} into the safety problem and an action-constrained optimal control problem. We consider the finite-horizon and the infinite-horizon \textit{worst-case safety value functions}, derived in Hamilton-Jacobi (HJ) reachability analysis \cite{bansal, Wabersich2023}, defined by
\begin{align}
        V(x, t)  := \max_{\ctrl(\cdot)} \min_{s \in [0, t]}    \constraintfcn(\traj(s)),\quad \quad
    V(x)  := \lim_{t \rightarrow\infty} V(x, t). \label{eq:safe-value}
\end{align}
Then, the maximal control invariant set within the constraint set $\constraintset$, denoted as $\safeset$, is characterized as the zero-superlevel set of the value function, $\safeset =\{x \mid V(x) \ge 0 \}$ \cite{fisac2018general}. If a trajectory exits the set $\safeset$, it inevitably violates the constraint \eqref{eq:cop-constraint}, i.e., $\traj(t) \notin \constraintset$ at some time $t > 0$. The set of all safe control inputs that prevent the safety violation, denoted as the \textit{maximal safe action set} (at each state), is determined by the value function as

\begin{equation}
\label{eq:opt-safe-action}
\mathcal{U}_{\mathrm{safe}}^*(x)
=
\begin{cases}
U,
& V(x) > 0, \\[2mm]
\left\{
u \in U
\;\middle|\;
\nabla V(x)^\top f(x,u) \ge 0
\right\},
& V(x) = 0.
\end{cases}
\end{equation}

Based on $\safeactionsetopt$, Problem \eqref{eq:constrained-opt} can be equivalently written as the following action-constrained form \cite{hao2024}: \vspace{-0.5em}
\begin{equation}
\label{eq:constrained-opt-act}
\inf_{\ctrl(\cdot) \in \mathcal{U}} \int_{0}^{\infty} e^{-\gamma t} \costfcn(\traj(t), \ctrl(t)) dt \quad \text{s.t.} \quad  \ctrl(t) \in \safeactionsetopt(\traj(t)) \;\; \forall t \ge 0.
\end{equation}
In practice, we can achieve a conservative solution of \eqref{eq:constrained-opt-act} by considering a more conservative safe action set, satisfying $\safeactionset(x) \subseteq \safeactionsetopt(x)$ for all $x \in \safeset$. In the classic safe control literature, these safe action sets are constructed with various methods such as CBFs or predictive control \cite{Ames2016, wabersich2021predictive, Wabersich2023}.

Finally, \eqref{eq:constrained-opt-act} can be further conservatively approximated using the safety filtering architecture \cite{Wabersich2023}:
\hruleopen
\noindent Reference Policy: \vspace{-0.5em}
\begin{equation}
\label{eq:reference-policy}
\piref = \min_{\pi \in \Pi} \int_{0}^{\infty} e^{-\gamma t} \costfcn(\traj(t), \ctrl(t)) dt \quad \text{s.t.} \quad \ctrl(t) = \pi(\traj(t)) \;\; \forall t \ge 0,
\end{equation}
\noindent Safety filter: \vspace{-0.5em}
\begin{subequations}
\label{eq:safety-filtering}
\begin{align}
\pisafe(\traj(t), \piref(\traj(t)))
& =
\arg\min_{u \in U}
\|u-\piref(\traj(t))\|^2 \label{eq:min-norm}
\\
\text{s.t.}
& \quad
u \in \safeactionset(\traj(t)).
\end{align}
\end{subequations}
\hruleclose
In \eqref{eq:reference-policy}, a \textit{reference policy} $\piref$ which primarily considers the task objective is optimized. Next, in the safety filter \eqref{eq:safety-filtering}, the filtered control input is determined by a mapping $\pisafe: \R^n \times U \rightarrow U$, which maps the state and an unfiltered action into a filtered safe action. Equation \eqref{eq:min-norm} considers the min-norm objective, which modifies the reference action only when it is necessary for safety. Other objectives can also be considered under various formulations, such as in robust or predictive variants. The filtered output can be equivalently represented as the addition between the unfiltered reference action $u$ and the filter correction term $\Delta\pisafe(x, u)$:
\begin{equation}
    \pisafe(x, u) = u + \Delta\pisafe(x, u).
    \label{eq:correction}
\end{equation}
In the CBF literature, $\Delta\pisafe(x, u)$ is also called the CBF augmentation input \cite{lavretsky2025servo}.

\textbf{Discriminating Hyperplane.} For control-affine dynamics, which encapsulates rigid-body robot dynamics, the safe action set can be effectively represented by the following halfspace constraint at each state, \vspace{-0.25em}
\begin{equation}
    \safeactionset(x) = \{u \in U \; | \; a(x)^\top u \ge b(x) \}. \label{eq:dh}
\end{equation}
Here, the pair $(a(x),b(x))$ defines a \emph{discriminating hyperplane} (DH) in the control input space, separating certified safe inputs from potentially unsafe inputs at each state \cite{l4dc}. This formulation provides a universal representation of safety filter constraints for control-affine dynamics, which includes classical CBF and HJ reachability (e.g., \eqref{eq:opt-safe-action}) filters as special cases. 
The key advantage in using this representation is that the safety filter can now be learned from data by directly learning the hyperplane parameters $(a_\theta(x),b_\theta(x))$, without requiring an explicit dynamics model or an explicit computation of safety certificates such as the safety value function $V$ in \eqref{eq:safe-value}. 
With the discriminating hyperplane, the min-norm filter correction is determined as $\Delta\pisafe(x, u) = \max\left\{0,  (b(x) - a(x)^\top u)a(x) / \|a(x)\|_2^2\right\}$, which is zero when $u$ already satisfies \eqref{eq:dh}.

\vspace{-0.5em}

\begin{figure}[t]
\centering
\vspace{-0.5em}
\includegraphics[width=\textwidth]{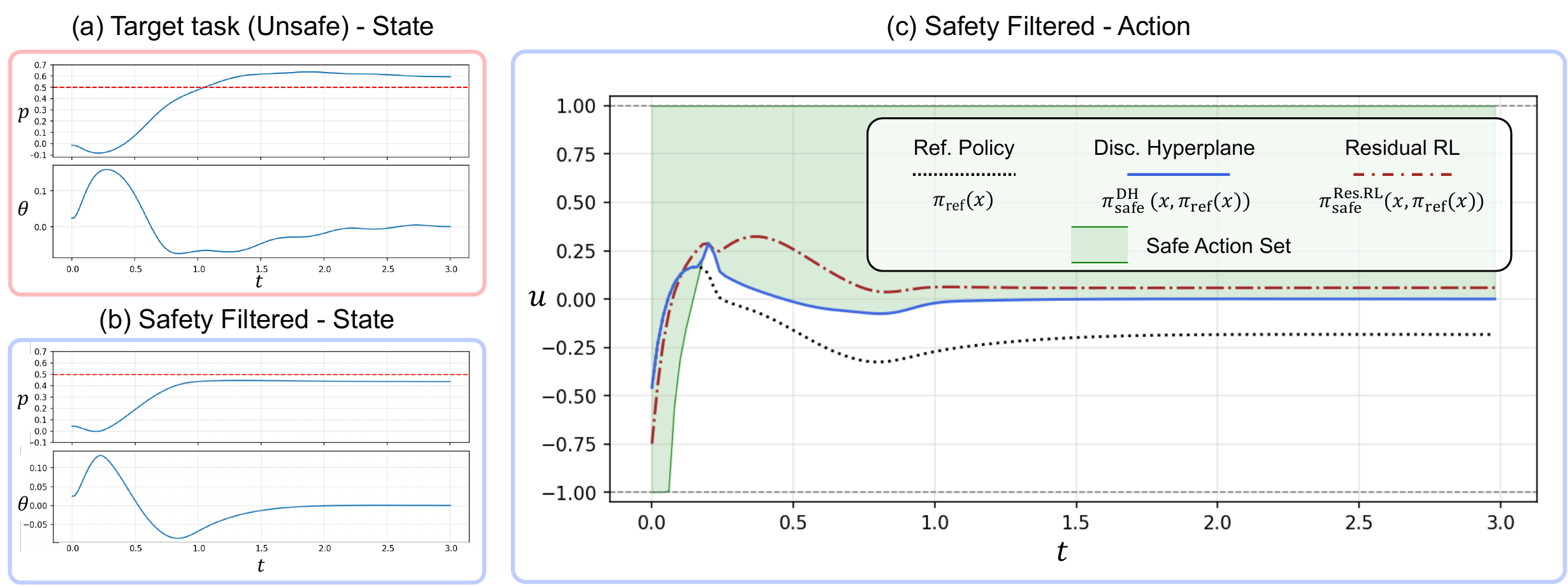}
\vspace{-1.25em}
 \caption{\small Cart-Pole Running Example. \textbf{(a)} Reference policy is trained for a task positioning the cart position to $p = 0.6$, and violates the safety constraint, $p \le 0.5$. \textbf{(b)} Learned safety filters can successfully prevent the state from safety violations. \textbf{(c)} $\pisafe$ under the learned min-norm discriminating hyperplane safety filter (blue) and residual RL (proposed method, brown). 
 }
\label{fig:running_example}
\vspace{-1em}
\end{figure}

\paragraph{Running Example (Cart-Pole).} We consider a simple cart-pole dynamics, with the state $x=[p;\theta;\dot{p};\dot{\theta}]$, where $p$ is the cart position and $\theta$ is the pole angle, and one-dimensional control input $u$ representing the torque applied to the pole, satisfying the bound $[-1, 1]$. The task is to stabilize the cart position to $p = 0.6$, while keeping the pole upright ($\theta = 0$). We consider the safety constraint of keeping the cart position $p \le 0.5$, which is intentionally chosen to conflict with the task objective for the illustrative purpose of the example. A simple MLP reference policy $\piref$ is learned through PPO \cite{schulman2017ppo} which achieves the task (Fig.~\ref{fig:running_example}.(a)). Then, a discriminating hyperplane learned through PPO in \cite{l4dc} is used to safety filter the reference policy (Fig.~\ref{fig:running_example}.(b)). The learned safe action set $\safeactionset$ is determined as $[-1, \frac{b(x)}{a(x)}]$ if $a(x) < 0$, and $[\frac{b(x)}{a(x)}, 1]$ if $a(x) > 0$. In Fig.~\ref{fig:running_example}.(c), the filtered action (blue) and the safe action set (green background) are visualized along the safe trajectory.

Although the discriminating hyperplane works effectively for the running example, since it does not directly learn the filtered safe control action but rather learns through the constraint representation, it generalizes less effectively for higher-dimensional systems. Also, it imposes the half-space constraint structure to $\safeactionset$, which might be impractical for more complex systems (e.g., beyond control-affine dynamics) whose safe action sets may be less structured (e.g, nonconvex and disconnected). In the next section, we propose an alternative approach of learning directly the correction term $\Delta \pisafe(x, u)$ through residual RL.

\vspace{-1em}
\section{Learning Safety Filtering with Residual Reinforcement Learning}
\vspace{-0.5em}

\subsection{Learning Safety Filtering with Residual RL}
\vspace{-0.5em}

To establish an RL-based framework and to deal with more complex systems like humanoids whose dynamics are inherently more uncertain and stochastic, we now consider an MDP-based probabilistic notion of safety. That is, we aim to learn $\safeactionset$ such that under $a_t \in \safeactionset (s_t)$, where $(s_t, a_t)$ is the discrete-time state and action pair at timestep $t$, we achieve the highest probability of the safety constraint satisfaction, which we call the \textit{safety rate}, given as $\Pr\big[s_t \in \constraintset \;\; \forall t\in[0, T]\big]$, where $T$ is the task time horizon. This is a stochastic reachability problem \cite{jang2025eigensafe}, and can be formulated as the \textit{maximal survival time problem}. As in \cite{l4dc}, if we learned the safe action set $\safeactionset$ with RL, we can learn it by considering $\max_{\safeactionset} \; \mathbb{E}_{a_t \sim \safeactionset(s_t)}\big[T_{\mathrm{surv}} \big]$, where $T_{\mathrm{surv}}$ is the time at which an episode terminates due to a constraint violation ($s_t \notin C$) or completion (in which $T_{\mathrm{surv}}\!=\!T$). 

The main benefit of solving the safety problem in this way, compared to existing approaches that solve the reachability value functions \eqref{eq:safe-value} with RL \cite{fisac2019bridging, huh2020safe, hsu2023isaacs, so2024train, li2025certifiable, yang2025cbf, oh2025safety, li2025verifiable,  kim2026deep, sharpless2026bellman}, is that the problem can be reformulated as a sum-over-reward problem given by
$\max_{\safeactionset} \; \mathbb{E}\big[\sum_{t=0}^{T_{\mathrm{surv}}} \mathbf{1}\big]$. In contrast, existing approaches typically involve a min-over-time objective, as in \eqref{eq:safe-value}. This fundamentally leads to a different class of Bellman equations \cite{fisac2019bridging}, for which off-the-shelf RL algorithms such as PPO are not directly applicable without specialized modifications.

With residual RL, we do not learn $\safeactionset$ but
intend to directly learn the correction term $\Delta\pisafe$ in \eqref{eq:correction}. Thus, $\Delta\pisafe$ is represented as a residual RL policy, whose inputs are the state $s_t$, and the (unfiltered) control action determined by the reference policy, $\piref(s_t)$. Also, it involves learning the min-norm filtering objective in \eqref{eq:min-norm}. Thus, the final form of our learning becomes
\begin{equation}
\max_{\Delta\pisafe^\theta} \; \mathbb{E}_{a_t \sim \pisafe^\theta(s_t, \piref(s_t))}\;\Big[\sum_{t=0}^{T_{\mathrm{surv}}} \mathbf{1} - \lambda ||\Delta\pisafe^\theta (s_t,a_t)  ||_2^{2}\Big],
\label{eq:residual-rl-obj}
\end{equation}
where $\pisafe^\theta(s_t, \piref(s_t)) = \piref(s_t) + \Delta\pisafe^\theta(s_t, \piref(s_t))$, and $\lambda$ is a weight of the min-norm objective penalty. The constant term $\mathbf{1}$ is known as the ``alive bonus'', and is included in many practical RL-based robot learning frameworks as one of the hand-designed reward terms \cite{he2024agile}. If the state $s_t$ is not directly accessible, we can replace it with observations $o_t$.

\textbf{Running Example (Continued).} We learn the residual RL safety filter for the cart-pole running example using PPO with the reward defined in \eqref{eq:residual-rl-obj}. 
~As illustrated in Fig.~\ref{fig:running_example}.(c), the residual RL policy closely approximates the min-norm safety filtered control input determined by the discriminating hyperplane, with slight conservativeness. This additional conservativeness acts as a robustness margin and helps improve the safety rate for humanoid balance tasks in the next section. 

\vspace{-0.5em}
\subsection{Learning Safety Filters for Humanoid Balance}
\vspace{-0.5em}

The main challenge of applying the residual RL framework established in the previous section to humanoids is its high dimensionality. For Unitree G1, the state and input dimensions are $n=58$, $m=29$. To learn an effective safety filter which explores enough in training to secure robustness and generalizability, we need to carefully guide the training of residual RL (Fig.~\ref{fig:method}.(c)). 

\textbf{General Balance Motion Tracking Policy.}
We train the reference policy using the balance motion tracking pipeline established in \cite{zhang2025hub}. However, instead of training a policy to track only a single reference motion, we generate a large and diverse set of balance motions using the method in \cite{pan2025agility}. In total, we generate 10,000 balance motions that cover a wide range of feasible humanoid configurations and behaviors. Each motion has varying level of difficulty, which we can empirically evaluate after the training of reference policy. Training on this diverse motion dataset encourages the safety policy to learn a more general balance prior, rather than overfitting to a specific trajectory.


\textbf{Reward Design for Residual RL.}
The residual policy is trained with a compact reward function that encourages balance recovery while preserving the behavior of the nominal controller. In addition to the two terms in \eqref{eq:residual-rl-obj}, we add two additional safety rewards: base velocity stabilization term to discourage excessive horizontal drift, and action smoothness penalty to reduce abrupt changes in control. This can be considered as additional objective terms in safety filtering. The detailed reward expressions and weights are provided in Appendix~\ref{app:reward}.

\textbf{Training Strategy.}
We train the residual policy using PPO \cite{schulman2017ppo} with an asymmetric actor-critic framework \cite{DBLP:conf/rss/PintoAWZA18}. The actor represents the deployable residual policy and only receives observations available at deployment time, including proprioceptive states and the reference policy action. In contrast, the critic is used only during training and can access privileged simulation information, such as ground-truth body states. 
Moreover, during training, two additional techniques are introduced to enhance the safety and robustness of the residual policy. First, we introduce a varying level of payloads at various locations of the robot body, and random pushes to the robot to expose it to failure or near-failure states for exploration. Without such excitation, the policy tends to learn a very conservative safety filter which is impractical. Next, the residual policy learns safety corrections for multiple reference policy checkpoints, which allows it to generalize against unseen unfiltered actions. Additional details including training parameters are provided in Appendix \ref{app:state space} and \ref{app:training parameters}.

\section{Experimental Results}

In this section, we evaluate the effectiveness of the proposed residual policy for improving balance-motion tracking under payload disturbances in both simulation and real-world hardware experiments. The central hypotheses we want to evaluate are
\begin{enumerate}[itemsep=0em, leftmargin=1.5em]
    \item Does decoupling the learning of safety and performance achieve a better performance--safety--robustness trade-off compared to single-policy training that jointly optimizes all objectives through a unified reward function?
    \item Does the proposed residual RL safety filtering framework generalize across diverse motions and previously unseen reference policies?
    \item Does the proposed residual RL safety filter transfer effectively to real-world hardware?
\end{enumerate}

\subsection{Experimental Setup}

We evaluate our method on the Unitree G1 humanoid robot across a range of balance tasks and payload disturbances in both simulator and real-world hardware. Simulation experiments are conducted using the IsaacGym \cite{NEURIP} simulator. 

We compare our method against three baselines: 1) the reference motion-tracking policy trained without payload domain randomization (denoted as Baseline), 2) the reference policy trained directly with payload domain randomization (denoted as Baseline*), and 3) the discriminating hyperplane safety filter trained with PPO \cite{l4dc} (denoted as DH). Note that the discriminating hyperplane demonstrates superior safety against classic safe RL methods in \cite{ray2019benchmarking}. We use the local errors relative to the robot base frame $E_{\mathrm{pos}\text{-}l}$ to measure tracking accuracy, and the fall rate to measure robustness and safety.

\subsection{Simulation Results}
For the Performance–Robustness Trade-off and Performance–Safety Trade-off evaluations, all policies are trained on a single reference motion. For the generalization evaluation, we train the policy on the entire balance dataset.

\begin{table}[t]
\setlength{\tabcolsep}{2pt}
\centering
\small 
\begin{minipage}[t]{0.48\linewidth}
\centering
\caption{Performance--Robustness Trade-off.}
\label{tab:performance-robustness} 
    \label{tab:performance-robustness}
    \begin{tabular}{lccc}
        \toprule
        \textbf{Policy} 
        & \makecell{\textbf{$\boldsymbol{E_{\mathrm{pos}\text{-}l}}$ $\downarrow$}\\\textbf{(w. payload)}} 
        & \makecell{\textbf{$\boldsymbol{E_{\mathrm{pos}\text{-}l}}$ $\downarrow$}\\\textbf{(w/o payload)}} 
        & \makecell{\textbf{Fall Rate$\downarrow$}\\\textbf{(w. payload)}} \\
        \midrule
        Baseline 
        & 38.6 mm 
        & 24.95 mm 
        & 7.29\% \\

        Baseline*
        & 43.28 mm 
        & 44.40  mm 
        & 1.20\% \\

           \textbf{ResSafe} 
        & 37.66 mm %
        & 34.04 mm 
        & 1.11\% \\
        \bottomrule
    \end{tabular}
\end{minipage}
\hfill
\begin{minipage}[t]{0.48\linewidth}
\setlength{\tabcolsep}{2pt}
    \centering
    \caption{Performance--Safety Trade-off.}
    \vspace{-0.8mm}
    \label{tab:performance--safety}
    \begin{tabular}{lcc}
        \toprule
        \textbf{Policy} (safety-cost weight) & \textbf{$\boldsymbol{E_{\mathrm{pos}\text{-}l}}$ $\downarrow$} & \textbf{Fall Rate$\downarrow$} \\
        \midrule
        Baseline* & 59.89 mm & 1.17\% \\
        Baseline* + Safety (50)& 101.68 mm & 0.00\% \\
        
        Baseline* + Safety (100)& 353.55 mm & 1.63\% \\
        \textbf{ResSafe} & 68.65 mm & 0.5\% \\
        \bottomrule
    \end{tabular}
\end{minipage}
\vspace{-1em}
\end{table}

\textbf{Performance--Robustness Trade-off.}
Table~\ref{tab:performance-robustness} reports the tracking error and fall rate of different policies. The baseline achieves reasonable tracking performance, but its fall rate increases under random payloads, indicating that accurate motion tracking alone does not guarantee robustness when the system dynamics change. Training the baseline directly with payload randomization improves robustness, but leads to higher tracking error. In comparison, our approach achieves the lowest fall rate while keeping the tracking error close to the nominal policy. This suggests that ResSafe provides a better balance between robustness and tracking performance.

\begin{figure}[t]
\centering
\vspace{-0.5em}
\includegraphics[width=1\textwidth]{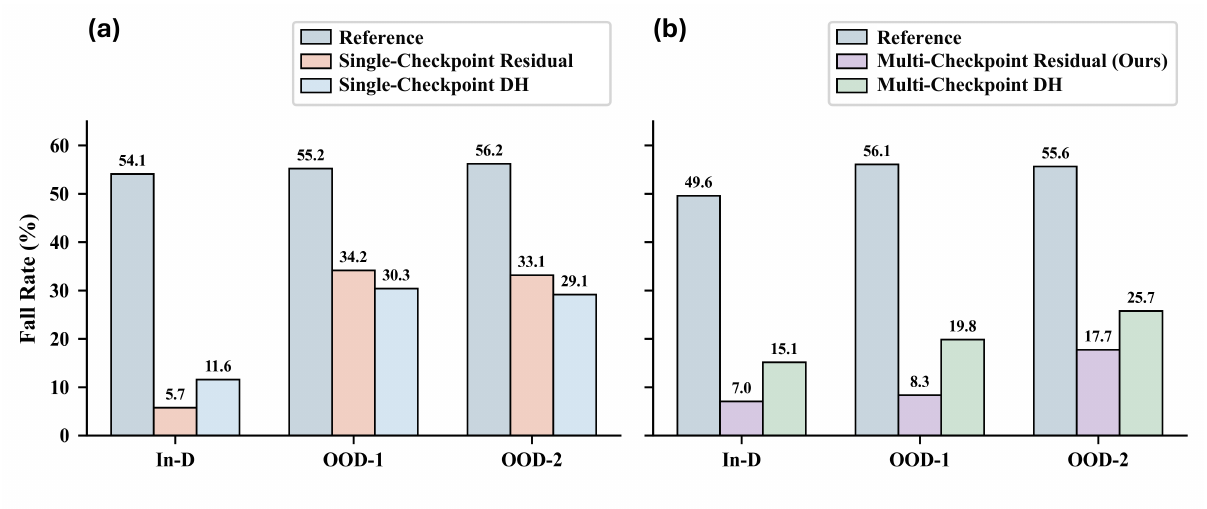}
\vspace{-2.5em}
 \caption{\small Generalization across different reference policies.
\textbf{(a)} Both the residual policy and the hyperplane policy are trained on a single reference-policy checkpoint. When they are applied to out-of-distribution reference-policy checkpoints, the fall rate increases sharply.
\textbf{(b)} Our method and the hyperplane policy are trained with half of the checkpoints in the dataset, which leads to a smaller safety degradation when evaluated with unseen checkpoints.}
\label{fig:generalization}
\vspace{-1em}
\end{figure}

\textbf{Performance--Safety Trade-off.}
Table~\ref{tab:performance--safety} further compares our method with policies trained using different safety-cost weights under random payloads. Adding a moderate safety cost can reduce the fall rate, but larger safety weights significantly degrade tracking performance and do not consistently improve safety. In contrast, ResSafe improves safety without causing severe tracking degradation. This shows the benefit of separating nominal motion tracking from residual safety correction.

\textbf{Generalization Across Different Reference Policies.}
We further evaluate whether the learned residual policy can generalize across different reference policies. As shown in Fig.~\ref{fig:generalization}.(a), when the residual policy is trained using a single reference policy checkpoint, its performance degrades significantly when applied to other checkpoints, resulting in a substantially higher fall rate. In contrast, a residual policy trained across multiple reference policy checkpoints maintains robustness when deployed with unseen reference policies (Fig.~\ref{fig:generalization}.(b)). This suggests that the learned residual correction captures general safety-relevant behaviors rather than overfitting to a particular reference controller. We also observe that the DH safety filter is inferior in both safety and robustness.

\subsection{Hardware Results}
\vspace{-0.5em}
\label{sec:hardware}

We further validate the proposed method on a physical humanoid robot. We train a policy on the entire balance dataset in simulation and deploy it directly on the real robot. The policies are tested for multiple reference motions with and without payloads attached to the limbs (Fig.~\ref{fig:frontfigure}). As shown in Table~\ref{tab:real_world_results}, 
our method outperforms both the baseline and the discriminating hyperplane in terms of fall rate and tracking error. Baseline*, achieves better tracking under payload because it is explicitly trained with a tracking reward; however, its tracking error is worse without payload across multiple motions  due to over-conservatism from domain randomization.
These results demonstrate that the residual safety filtering policy can be transferred to the real robot and improve safety and robustness when the robot performs challenging motion tasks.


\vspace{-0.5em}
\section{Conclusion}
\label{sec:conclusion}
\vspace{-0.5em}

We presented ResSafe, a performance--safety decoupling framework that combines a task-oriented reference policy with a residual RL safety filter. We showed that the learned residual policy acts similarly to a min-norm safety filter correction and demonstrated that this decoupled learning paradigm achieves a superior performance--safety--robustness trade-off compared to conventional single-policy training approaches. These results highlight the potential of decoupling performance and safety as a pathway toward reusable and generalizable safety mechanisms for humanoids.

\vspace{-0.5em}
\section{Limitations}
\vspace{-0.5em}

While the proposed framework demonstrates promising empirical performance, several limitations remain. First, because the safety filter is learned from data, it does not provide hard safety guarantees in the sense of classical safety filters. Second, the residual policy consistently exhibits more conservative behavior than the discriminating hyperplane filter, and a theoretical understanding of this phenomenon remains an open question. Third, our approach relies on standard sim-to-real techniques used in humanoid RL, such as system identification and domain randomization, and does not explicitly address sim-to-real transfer, which contributes to the gap observed between simulation and hardware experiments. Fourth, the generalization capability of the learned safety filter is limited by the diversity of the training motion dataset, although fine-tuning can be used to adapt to new motions. Finally, the current framework operates as a blind safety filter without exteroceptive perception. In contrast, human's balancing and safety behaviors rely heavily on visual feedback and environmental awareness. Incorporating perception and leveraging environmental affordances for safety, inspired by recent work such as \cite{He_2026_CVPR}, is an important direction for future research.

\begin{table}[t]
\centering
\vspace{-1em}
\caption{Real-World Results (\# indicates motion id in dataset).}
\label{tab:real_world_results}
\setlength{\tabcolsep}{2.5pt}
\renewcommand{\arraystretch}{1.05}
\begin{tabular}{lcc@{\hspace{6pt}}|@{\hspace{6pt}}lcc}
\toprule
\textbf{Method} & \textbf{Fall} & $\boldsymbol{E_{\mathrm{pos}\text{-}l}}$\,(mm)
& \textbf{Method} & \textbf{Fall} & $\boldsymbol{E_{\mathrm{pos}\text{-}l}}$\,(mm) \\
\midrule
\rowcolor{gray!20}
\multicolumn{3}{l}{\textbf{(a) \#9350 (w/ payload)}}
& \multicolumn{3}{l}{\textbf{(b) \#1665 (w/o payload)}} \\
Baseline  & 10/10 & 84.13 $\pm$ 12.99 & Baseline  & 8/10 & 80.42 $\pm$ 23.16 \\
Baseline* & 0/10  & 47.73 $\pm$ 4.01  & Baseline* & 2/10 & 73.67 $\pm$ 13.53 \\
DH        & 1/10  & 67.88 $\pm$ 20.13 & DH        & 3/10 & 87.95 $\pm$ 21.10 \\
\textbf{Ours} & 2/10 & 49.81 $\pm$ 11.65
              & \textbf{Ours} & \textbf{1/10} & 76.43 $\pm$ 6.22 \\
\midrule
\rowcolor{gray!20}
\multicolumn{3}{l}{\textbf{(c) \#6337 (w/ payload)}}
& \multicolumn{3}{l}{\textbf{(d) \#8407 (w/ payload)}} \\
Baseline  & 10/10 & 73.56 $\pm$ 14.68 & Baseline  & 5/10 & 75.49 $\pm$ 28.52 \\
Baseline* & 5/10  & 62.00 $\pm$ 14.43 & Baseline* & 2/10 & 48.49 $\pm$ 13.62 \\
DH        & 9/10  & 82.62 $\pm$ 22.98 & DH        & 4/10 & 66.10 $\pm$ 15.04 \\
\textbf{Ours} & \textbf{0/10} & 70.01 $\pm$ 3.11
              & \textbf{Ours} & \textbf{0/10} & 61.01 $\pm$ 4.41 \\
\bottomrule
\end{tabular}
\vspace{-1em}
\end{table}

\acknowledgments{We thank Siming He, Christopher Strong, Qiayuan Liao, and Junfeng Long for insightful discussions, and Chuye Hong, Yiyang Shao, and Haoyi Niu for their help with the hardware experiments, and VESSL AI  for supporting this research with GPU compute resources. This work was supported in part by the NSF Safe Learning Enabled Systems Program, Design of Robustly Implementable Autonomous and Intelligent Machines, DARPA award number HR00112490425, and the UCLA Samueli School of Engineering.}


\bibliography{reference}  

\newpage
\appendix
\section{Residual Reinforcement Learning Details}
\subsection{Reward Design}
\label{app:reward}

This section summarizes the reward design of ResSafe, which is decoupled from the tracking reward.

\begin{center}
\captionof{table}{Reward components and weights.}
\label{tab:reward_design}
\setlength{\tabcolsep}{4pt}
\renewcommand{\arraystretch}{1.08}
\resizebox{0.8\textwidth}{!}{
\begin{tabular}{lccc}
\toprule
\textbf{Term} & \textbf{Expression} & \textbf{Weight} & \textbf{Remarks} \\
\midrule

\multicolumn{4}{c}{\textbf{Reward}} \\
\midrule

Alive bonus
& $1$
& $1.6$
&  \\


Base linear velocity stabilization
& $\exp\!\left(-\dfrac{\|v_{xy,t}^{\text{base}}\|_2^2}{\sigma_{\text{lin}}^2}\right)$
& $2.5$
& $\sigma_{\text{lin}} = 0.3$ \\

Base angular velocity stabilization
& $\exp\!\left(-\dfrac{\|\omega_{xyz,t}^{\text{base}}\|_2^2}{\sigma_{\text{ang}}^2}\right)$
& $4$
& $\sigma_{\text{ang}} = 1.2$ \\

\midrule
\multicolumn{4}{c}{\textbf{Penalty}} \\
\midrule

Termination
& $1$
& $-60$
&  \\

Minimal intervention
& $\left\|a_t^{\text{applied}} - a_t^{\text{reference}}\right\|_2^2$
& $-0.07$
&  \\

Action smoothness 
& $\left\|a_t^{\text{applied}} - a_{t-1}^{\text{applied}}\right\|_2^2$
& $-0.01$
&  \\

\bottomrule
\end{tabular}
}
\end{center}

\subsection{State Space Design}
\label{app:state space}
This subsection describes the state-space design used in our asymmetric actor--critic residual reinforcement learning.

\begin{center}
\captionof{table}{State space information of the critic network input.}
\label{tab:critic_obs_reduced}
\begin{tabular}{lc}
\toprule
\textbf{State Term} & \textbf{Dimensions} \\
\midrule
Rigid body position & 87 \\
Rigid body rotation & 180 \\
Rigid body velocity & 90 \\
Rigid body angular velocity & 90 \\
Applied action & 29 \\
Reference policy action & 29 \\
\midrule
\textbf{History State Term} & \textbf{Dimensions} \\
\midrule
DoF position & 29 \\
DoF velocity & 29 \\
Base angular velocity & 3 \\
Projected gravity & 3 \\
Applied action & 29 \\
\midrule
\textbf{History steps} & \textbf{2} \\
\midrule
\textbf{Total dim} & $\mathbf{505 + 93\times 2 = 691}$ \\
\bottomrule
\end{tabular}
\end{center}

\begin{center}
\captionof{table}{State space information of the actor network input.}
\label{tab:actor_obs_reduced}
\begin{tabular}{lc}
\toprule
\textbf{State Term} & \textbf{Dimensions} \\
\midrule
DoF position & 29 \\
DoF velocity & 29 \\
Base angular velocity & 3 \\
Projected gravity & 3 \\
Applied action & 29 \\
Reference policy action & 29 \\
\midrule
\textbf{Total dim} & $\mathbf{122}$ \\
\bottomrule
\end{tabular}
\end{center}

\subsection{Training Parameters}
\label{app:training parameters}
In this subsection, we provide the training parameters used for ResSafe.

\begin{center}
\captionof{table}{Hyperparameters.}
\label{tab:hyperparameters}
\begin{tabular}{lc}
    \toprule
    \textbf{Hyperparameter} & \textbf{Value} \\
    \midrule
    Optimizer & Adam \\
    Adam coefficients $(\beta_1, \beta_2)$ & $(0.9, 0.999)$ \\
    Learning rate & $1 \times 10^{-3}$ \\
    Rollout horizon $N_{\mathrm{steps}}$ & $36$ \\
    Number of mini-batches & $12$ \\
    Discount factor $\gamma$ & $0.99$ \\
    PPO clipping parameter & $0.2$ \\
    Entropy coefficient & $0.001$ \\
    Gradient clipping norm & $5$ \\
    Value loss coefficient & $1$ \\
    Initial policy std. & $0.5$ \\
    Number of learning epochs & $5$ \\
    MLP hidden layers & $[512,\ 256,\ 128]$ \\
    \bottomrule
\end{tabular}
\end{center}

\subsection{Domain Randomization}

We use the same dynamics domain randomization as in HuB~\cite{zhang2025hub}, while additionally introducing payload perturbations and a curriculum learning scheme for random push disturbances.

\begin{center}
\captionof{table}{Domain-randomization settings.}
\label{tab:disturbance_settings}
\setlength{\tabcolsep}{8pt}
\renewcommand{\arraystretch}{1.15}
\resizebox{0.8\textwidth}{!}{
\begin{tabular}{cc}
\toprule
\textbf{Term} & \textbf{Value} \\

\midrule

\multicolumn{2}{c}{\textbf{Push Disturbance Curriculum Learning}} \\
\midrule

Curriculum variable
& $v_{xy}\in\mathcal{U}(0,v_{\max})\,\mathrm{m/s}, 
\quad \mathrm{interval}=4\,\mathrm{s}$ \\
Initial value
& $v_{\max}=0.0\,\mathrm{m/s}$ \\
Maximum value
& $v_{\max}=0.5\,\mathrm{m/s}$ \\
Update rule
& $
\begin{aligned}
v_{\max} &\leftarrow \min(v_{\max}+0.01,\ 0.5), 
&& \text{if episode length}>300, \\
v_{\max} &\leftarrow \max(v_{\max}-0.01,\ 0.0), 
&& \text{if episode length}<250, \\
v_{\max} &\leftarrow v_{\max}, 
&& \text{otherwise}.
\end{aligned}
$\\
\midrule
\multicolumn{2}{c}{\textbf{Payload Disturbance}} \\
\midrule
Payload links
& $\{\text{L/R wrist},\ \text{L/R elbow},\ \text{L/R shoulder}\}$ \\

Added mass per link
& $m_{\mathrm{add}}\sim\mathcal{U}(0,3)\,\mathrm{kg}$ \\

\bottomrule
\end{tabular}
}
\end{center}

\section{Experimental Details}
\label{app:experiment_details}

We provide additional implementation details for the experiments reported in the main paper.

\subsection{Performance--Robustness Trade-off}

The performance--robustness trade-off experiments are conducted in a single motion setting, in which all methods are trained and evaluated on the same reference motion. In HuB, the final policy is obtained through teacher--student distillation. We use the student reference policy to train our general residual policy. However, in the performance--safety trade-off experiment in Section~\ref{ps}, we aim to incorporate an additional safety reward during training. To avoid introducing distillation as an additional confounding factor, we use the teacher policy as the reference policy in both trade-off experiments. The residual policy training uses the state space shown in Table~\ref{tab:critic_obs_reduced} for both the actor and critic networks.

\subsection{Performance--Safety Trade-off}
\label{ps}

Similarly, the performance--safety trade-off experiments are conducted in a single motion setting, in which all methods are trained and evaluated on the same reference motion. We use the teacher policy as the reference policy, and the residual policy training uses the state space shown in Table~\ref{tab:critic_obs_reduced} for both the actor and critic networks.

\subsection{Generalization Across Different Reference Policies}

In contrast, in the generalization experiment, the residual policy is trained on the full motion dataset using the student general reference policy, since the goal is to evaluate whether the learned residual policy can transfer across different reference policy checkpoints and a broader motion distribution.

\subsection{Hardware Results}

In the hardware experiment, the residual policy is trained on the full motion dataset using the student general reference policy and evaluated on different motions from the dataset.

\end{document}